\documentclass[letterpaper, 10 pt, conference]{ieeeconf}  % Comment this line out if you need a4paper

\IEEEoverridecommandlockouts                              % This command is only needed if 
\usepackage{cite}
\usepackage{amssymb}
\usepackage{amsmath}
\usepackage[table]{xcolor}
\usepackage{multirow}
\usepackage{graphicx}
\usepackage{xspace}
\usepackage{subcaption}
\usepackage{caption}
\usepackage{floatrow}
\usepackage[labelformat=simple]{subcaption}	
\DeclareMathOperator*{\argminA}{arg\,min} % Jan Hlavacek
\usepackage[normalem]{ulem}

\newcommand{\ctog}{cost-to-go\xspace}
\newcommand{\cspace}{C-space\xspace}

\newcommand{\ctogplanner}{c2g-network planner\xspace}
\title{Learning Continuous Cost-to-Go Functions for Non-holonomic Systems}
\author{Jinwook Huh, Daniel D. Lee and Volkan Isler% <-this % stops a space
\thanks{All authors are with the Samsung AI Center NY, 123 West 18th Street, New York, New York 10011}%
}
\date{March 2021}

\begin{document}

\maketitle

%to IROS, Papers should be  six pages in length,  with up to two extra pages (180EUR charge per extra page). Note: The maximum number of pages includes references. EG: if the paper's technical section is 6-pages, then one can include up to 2-pages of additional info e.g. references, for the stated cost (i.e. 180EUR/pages * 2 pages = 360EUR extra cost). A short video (1-minute maximum) can also be attached to your paper to supplement results (see Instructions below). The IROS  Conference Paper Review Board  (like ICRA’s CEB) rigorously reviews all submissions and then through the IROS 2021 Senior Program Committee to render acceptance decisions. Best Paper Awards in various categories are also selected by their respective committees. 

\begin{abstract}

This paper presents a supervised learning method to generate continuous cost-to-go functions of non-holonomic systems directly from the workspace description. Supervision from informative examples reduces training time and improves network performance.  
The manifold representing the optimal trajectories of a non-holonomic system has high-curvature regions which can not be efficiently captured with uniform sampling. To address this challenge, we present an adaptive sampling method which makes use of sampling based planners along with local, closed-form solutions to generate training samples. The cost-to-go function over a specific workspace is represented as a neural network whose weights are generated by a second, higher order network. The networks are trained in an end-to-end fashion. In our previous work, this architecture was shown to successfully learn to generate the cost-to-go functions of holonomic systems using uniform sampling. In this work, we show that uniform sampling fails for non-holonomic systems. However, with the proposed adaptive sampling methodology, our network can generate near-optimal trajectories for non-holonomic systems while avoiding obstacles. Experiments show that our method is two orders of magnitude faster compared to traditional approaches in cluttered environments.

% for the learning of networks in an end-to-end fashion from data obtained by traditional motion planners. The learned network based on this adaptive sampling successfully represents the cost in high-curvature regions of cost-to-go in the C-space of nonholonomic systems.
% Furthermore, the network generates continuous cost-to-go functions represented by another neural network, and we can use the cost-to-go function network for motion planning. The planners with the network generate near-optimal trajectories for non-holonomic systems while avoiding obstacles. 
% We demonstrate the performance based on the adaptive sampling method by comparisons with uniform sampling approaches.
% The experimental results show that ours is two orders of magnitude faster compared to traditional approaches in cluttered environments.
 
\end{abstract}

\section{Introduction}

Non-holonomic constraints of a system are non-integrable kinematic constraints involving the derivatives of the system's state variables. For example, a simple car, whose state can be represented as position and orientation on a plane, is subject to a turning constraint: it cannot instantaneously change its orientation to an arbitrary angle. Hence, the magnitude of the derivative of the orientation is upper-bounded. Non-holonomic constraints are inherent to the system's kinematics and cannot be removed using algebraic tricks. A classical example of a non-holonomic robotics system is the Dubin's car~\cite{dubins1957curves} which can go forward and change its orientation subject to a maximum curvature constraint.
This constraint can also be expressed as a lower bound on the turning radius. 
The Reeds-Shepp car~\cite{reeds1990optimal} is slightly more general: it can flip its motion direction (i.e. back up) but it must still satisfy the curvature constraint when moving.

It is possible to compute optimal solutions for non-holonomic systems for some special cases. For example, for the Dubins car, it was shown that the optimal solution between two configurations is composed of at most three segments where each segment is either a line segment or an arc of the unit circle where the unit corresponds to the minimum turning radius~\cite{dubins1957curves}. In other words, the car either goes straight or turns at maximum curvature. This structure allows for computing the optimal solution simply by enumerating combinatorially different paths and picking the best one (Once the combinatorial structure is fixed, the path itself can be computed using elementary geometry.) A similar solution structure is also known for the Reeds-Shepp car~\cite{reeds1990optimal}.

\begin{figure}
    \centering
    \includegraphics[height=3.9cm]{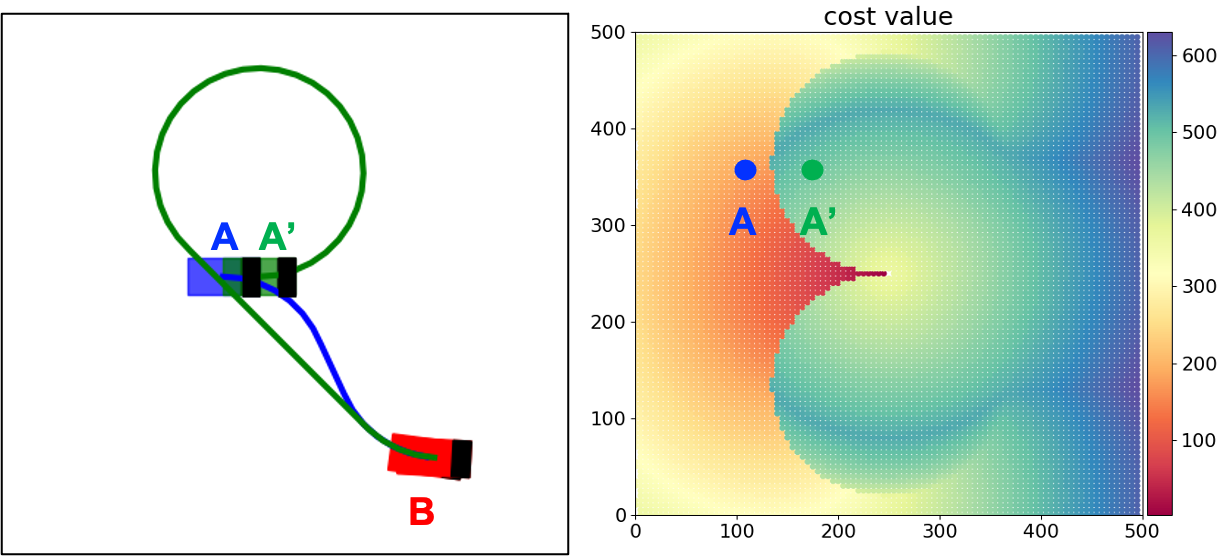}
    \caption{(Left) Configuration $A'$ is obtained by slightly perturbing $A$. The cost of $A'$ to the goal $B$ is drastically different from the cost between $A$ and $B$. (Right) $A$ and $A'$ are in the region with high curvature of cost-to-go.}
    \label{fig:high_curvature}
\end{figure}

Motion planning for non-holonomic systems in general is difficult. In the presence of obstacles, the optimal solution no longer satisfies the structural properties mentioned above. For example, even in the case of turning a corner with a Dubins car, turning with maximum curvature may not be optimal~\cite{koval2019turning}. To see why non-holonomic systems are challenging for commonly used sampling-based planners such as RRT and variants, it is helpful to visualize the space of allowable trajectories as a manifold where neighborhoods are formed by connecting nearby configurations satisfying the non-holonomic constraint. Now, consider the example in Fig.~\ref{fig:high_curvature} where the initial configuration $A$ and the goal configuration $B$ have the same orientation but different positions. The geodesic distance between these two configurations is relatively small as can be seen from the figure.
Next, imagine slightly perturbing $A$ to the configuration $A'$ as shown in the figure. The shortest path from $A'$ to $B$ is drastically longer. This means that configurations $A$, $A'$ and $B$ lie in a region with high curvature. Capturing the connectivity of such regions with sampling based planners is difficult because sampling a small number of ``nearby" points may not be sufficient to connect such local regions  \cite{lavalle2001randomized,kleinbort2018probabilistic,hauser2016asymptotically,li2016asymptotically}. %\vtxt{maybe we should add some references to kinodynamic planning}

In our recent work, we introduced a novel path planning approach where we use \ctog{} values obtained from a traditional planner to fit a continuous \ctog{} function over the configuration space (\cspace{}) represented as a neural network. We then showed that we can train a higher order network which can directly output this function from a representation (e.g. a picture) of the robot's workspace. In~\cite{huh2020cost} we showed that after training, our approach can be used to compute continuous trajectories almost instantaneously and generalizes well across workspaces. In the present work, we seek to extend this approach to non-holonomic systems.

Unfortunately, the existence of high-curvature regions on the motion planning manifold also makes it difficult for our approach to efficiently represent the cost-to-go function using uniform sampling. 

In this paper, we show how this difficulty can be overcome using an adaptive sampling approach. Specifically, we present an adaptive sampling methodology based on sampling according to the curvature of cost-to-go. The suggested approach allows the network to learn cost-to-go efficiently and to output continuous cost-to-go functions in C-space from a workspace input. The planner with the cost-to-go function network rapidly generates near-optimal trajectories satisfying constraints and avoiding obstacles in cluttered environments. We demonstrate that our approach is orders of magnitude faster than traditional approaches for trajectory generation in cluttered environments.

%  In supervised training of networks, informative examples reduce training time and improve network performance.  
% The manifold representing the valid trajectories of a non-holonomic system has high-curvature regions which can not be efficiently captured with uniform sampling. To address this challenge, we present an adaptive sampling method which combines sampling based planners with local, closed-form solution to generate training samples. The cost to go function over a specific workspace is represented as a neural network whose weights are generated by a second, higher order network. The networks are trained in an end-to-end fashion. In our previous work, this architecture was shown to successfully learn to generate the cost-to-go functions of holonomic systems using uniform sampling. In this work, we show that uniform sampling fails for non-holonomic systems. However, with the proposed adaptive sampling methodology, our network can solve  generate near-optimal trajectories for non-holonomic systems while avoiding obstacles. Experiments show that the our method is two orders of magnitude faster compared to traditional approaches in cluttered environments.
%\vtxt{jinwook please complete: a few sentences on our approach and results}

%We generate dataset from two phase RRT* construction for 
%sampling across vertices in Trees
%

%\input{old-intro}
\section{Related Work}

Many robotics applications necessitate motion planning to move a system from one configuration to another configuration while satisfying constraints and avoiding obstacles \cite{lynch2017modern,choset2005principles,lavalle2006planning}. Robotic systems such as car-like robots are subject to inherent non-holonomic constraints by kinematics. These constraints make motion planning more difficult since two feasible configurations may not be directly connected via a C-space shortest path due to kinematic constraints. 

% \begin{itemize}
%     \item Dimitry paper
%     \item soft constraints
% \end{itemize}

For a simple car model in free space, Dubins \cite{dubins1957curves} and Reeds and Shepp \cite{reeds1990optimal} showed that there the shortest path for the respective systems can be expressed as a combination of a small number of arc turning and straight line primitives. However, there is no analytical solution to the shortest path problem when such systems operate in cluttered environments. 

Solutions for non-holonomic motion planning in cluttered environments include traditional lattice-based planners which apply discrete search with a sequence of predefined control primitives~\cite{pivtoraiko2009differentially}. Rapidly-exploring Random Tree (RRT) planners solve  kinodynamic problems \cite{lavalle2001randomized} by randomly exploring on C-space with random continuous control inputs. However, in non-holonomic systems, they require exhaustive random samples and collision checks as well as post-processing for smoothing \cite{latombe1991robot,laumond1998robot}.
 
%  There are some research on navigation functions for the control of nonholonomic systems \cite{tanner2003nonholonomic,roussos20083d} and differential dynamic programming with nonholonomic constraints \cite{tassa2014control} generating state and control sequences minimizing a given cost function.
 
Although there are many approaches to the optimal trajectory for non-holonomic systems such as navigation functions \cite{tanner2003nonholonomic,roussos20083d} and differential dynamic programming \cite{tassa2014control}, these approaches are vulnerable to local minima and difficult to tune parameters in cluttered environments.

% The performance and result with navigation functions are sensitive for parameter tuning and it is hard for irregular obstacle shapes.

% There is differential dynamic programming with non-holonomic constraints \cite{tassa2014control} for trajectory optimization which solves state and control sequences by minimizing a given cost function. However, these approaches are vulnerable to local minima and hard to define cost functions in obstacle environments.

%However, these approaches doesn't consider obstacles in the environments and it is hard to define convex cost function in the obstacle environments.

Recently, deep neural networks have been proposed for motion planning. Tamar et al. suggest a value iteration network, which is a similar concept to learning cost-to-go. They do not address non-holonomic constraints \cite{tamar2016value}.
Motion planning network \cite{qureshi2018motion} outputs preliminary trajectories for traditional planners and it is extensible to the problem with non-holonomic constraints \cite{johnson2020dynamically}. In addition, there are some deep neural networks for good sample distributions  \cite{ichter2018learning,kumar2019lego,molina2020learn,ichter2020learned}. Zeng et al. \cite{zeng2019end} suggests a neural network which outputs cost volumes for computing costs of sample trajectories. The planner chooses a minimum cost trajectory among sample trajectories. These neural network approaches show performance improvement over traditional planners using modern deep neural networks. However, they use uniform samples from trajectories by traditional planners or human demonstrations for network training without considering sample efficiency.

% Recently, many deep neural networks are proposed for motion planning. Tamar et al. suggest value iteration network, which is similar concept to learning of cost-to-go without nonholonomic constraints \cite{tamar2016value}.
% Motion planning network \cite{qureshi2018motion} outputs preliminary trajectories for efficient sampling-based planners by encoding workspace pointclouds. Although this motion planning network extends to a problem with nonholonomic constraints \cite{johnson2020dynamically}, it still requires traditional planners to connect two local configurations generated by the network
% In addition, for efficient sampling-based planners, some deep neural networks learn to output good sample distributions based on trajectories by traditional planners \cite{ichter2018learning,kumar2019lego,molina2020learn,ichter2020learned}.
%and the network generates a target local configuration from local map.
%a sequence of control inputs makes sample complexity.

%Although there are many neural network approaches for motion planning, They use uniform sampling for training of neural networks. 

% There are some classical literature regarding sampling around surface of collision configurations for efficient tree exploration of sampling-based planners \cite{amato1998obprm,rodriguez2006obstacle,yeh2012uobprm}. %These approaches focus on efficient collision avoidance, not nonholonomic constraints.
%to output trajectory directly 
In general, training network for motion planning requires a massive dataset, since a trajectory is generated by a sequence of control inputs. Therefore, many previous neural network approaches exploit successful trajectories by traditional motion planners to create training datasets \cite{ichter2018learning,qureshi2018motion,johnson2020dynamically,molina2020learn,ichter2020learned}. 
However, all data samples are not equally important for the training of neural networks \cite{katharopoulos2018not,fan2017learning}. Adaptive sampling or boosting can accelerate training and improve performance. Although there are some classical approaches on adaptive sampling for sampling based planners \cite{amato1998obprm,rodriguez2006obstacle,yeh2012uobprm}, most neural network approaches for motion planning do not handle the issue of sampling to improve the performance of training. Our primary contribution to neural network based motion planning is a novel adaptive sampling method which captures high curvature regions of cost-to-go for the non-holonomic systems but using a combination of RRT based and optimal, closed-form local planners.

\section{Problem Statement}
% \vspace{-3mm}

%\subsection{Problem Definition}

Our representative non-holonomic system is a Reeds-Shepp car with a minimum turning radius $\rho$ \cite{laumond1998guidelines}. %\vtxt{ref.}
%We define a simple car model with a lower bounded turning radius $\rho$. 
Its configuration space is given by $\mathcal{C} = {\mathbb{R}}^2 \times {\mathbb{S}}^1$. A configuration $q \in \mathcal{C}$ is denoted as $ q = (x,y,\theta)$ and its time derivative $\dot{q} = (\dot{x}, \dot{y}, \dot{\theta})$. The system has a nonholonomic constraint such that $\dot{x}\cos{\theta} - \dot{y}\sin{\theta} = 0$. 
The control input $u$ has longitudinal velocity, steering angle, and gear direction for forward and backward directions. 

% \vtxt{don't we need to fix the goal to define cost to go?}
The cost-to-go function is a real-valued function $L(q)$ on the configuration space $q \in \mathcal{C}$. The optimal control input can be as satisfies,
$$u^{*}_k(q_{k}, q_{\text{goal}}) = \argminA_{u'} \{l(q_k, u') + L^{*}(q_{k+1}, q_{\text{goal}})\},$$
$$
q_{k+1} = g(q_k,u)
$$
where $l(q_k, u')$ is a constant cost given control input $u$, $q_{\text{goal}}$ is a goal configuration, $g$ is forward kinematics function of the system, and $L^{*}(q_{k+1})$ is the minimum cost-to-go value at $q_{k+1}$. In this paper, we define the cost of a collision-free path as the length of the path.

%and Backer and Kirkpatrick \cite{backer2007finding} suggest segments of collision-free Dubins path based on feature points in the environment.  
Although algorithms to generate optimal Dubins and Reeds-Shepp curves exist for environments with no obstacles, there is no analytical solution to the shortest path problem in cluttered environments. There are a few special cases to solve the problem. Agarwal et al. \cite{agarwal2002curvature} solve the shortest path problem inside polygonal obstacles. Koval and Isler~\cite{koval2019turning} show that the optimal solution does not necessarily turn at the maximum curvature. %Thus we consider a network to compute cost-to-go for Dubins or Reeds-Shepp cars in cluttered environments.
%This paper focuses on fast trajectory generation in cluttered environments by deep neural network approach based on the higher order function network.

%The cost-to-go function depends on workspace conditions and a goal configuration. We suggest a network to 
For a specific workspace $\mathcal{W}$,
the cost-to-go (c2g) function returns the cost between two configurations $f_{\mathcal{W}}: \mathcal{C}  \times \mathcal{C}  \rightarrow
[0, \infty)$.
%, where $\mathcal{W}$ is an input workspace (obstacle map).
Specifically, the function $f_{\mathcal{W}}$ is a cost-to-go function for any given configurations $s$ and $t$ in the configuration space $\mathcal{C}$, which returns the cost for the nonholonomic system to traverse a collision-free path while satisfying the nonholonomic constraint from $s$ to $t$. We train a network to output a function of cost-to-go instead of cost-to-go values on $\mathcal{C}$-space.

Since the cost-to-go is a result of a sequence of control inputs between two configurations and it is not a simple Euclidean distance between them, the cost-to-go function is not efficiently trained with a set of points distributed uniformly over the C-space. Therefore, we also focus on effective dataset generation for efficient training of the network.

%The cost-to-go value is changed according to the relative pose not relative distance. 
%In addition, collision avoidance depends on the configuration. It needs cover whole configuration space. so it requires a lot of samples.

%The cost-to-go function is relevant with control input. In other words, the cost-to-go values is defined by control inputs between two neighbors.

\section{Dataset Generation with Adaptive Sampling} \label{sec:methods}

\begin{figure}
\centering
\begin{subfigure}[b]{0.32\columnwidth}
\includegraphics[width=0.95\textwidth]{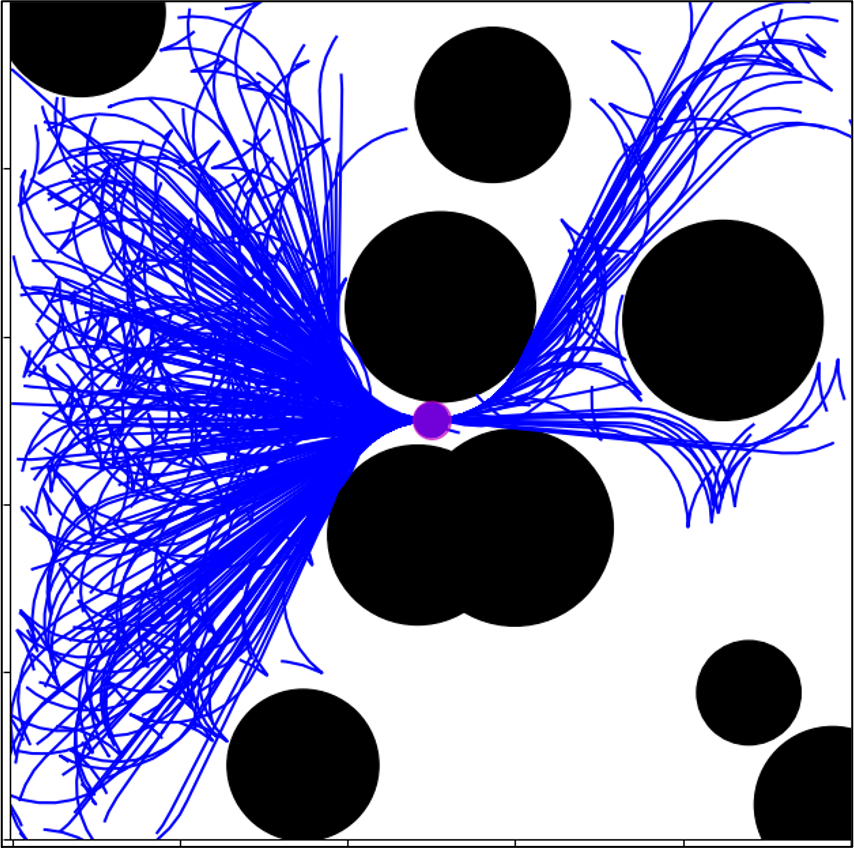}
\caption{} \label{fig:dataset_reeds_shepp}
\end{subfigure}
\begin{subfigure}[b]{0.32\columnwidth}
\includegraphics[width=0.95\textwidth]{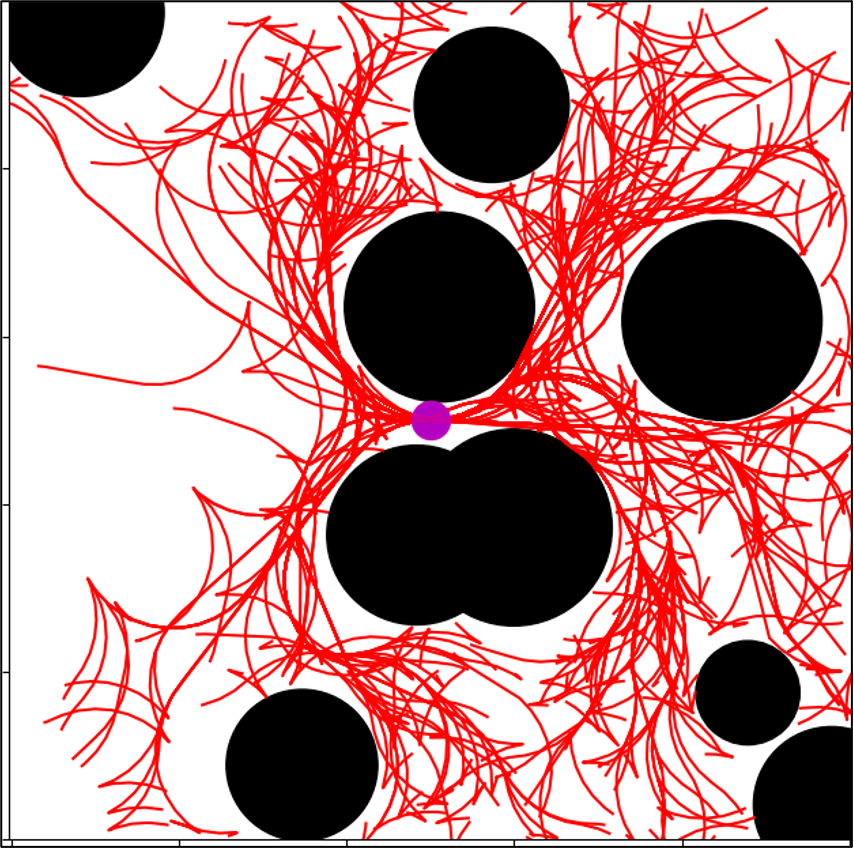}
\caption{} \label{fig:dataset_rrt_star}
\end{subfigure}
\begin{subfigure}[b]{0.32\columnwidth}
\includegraphics[width=0.95\textwidth]{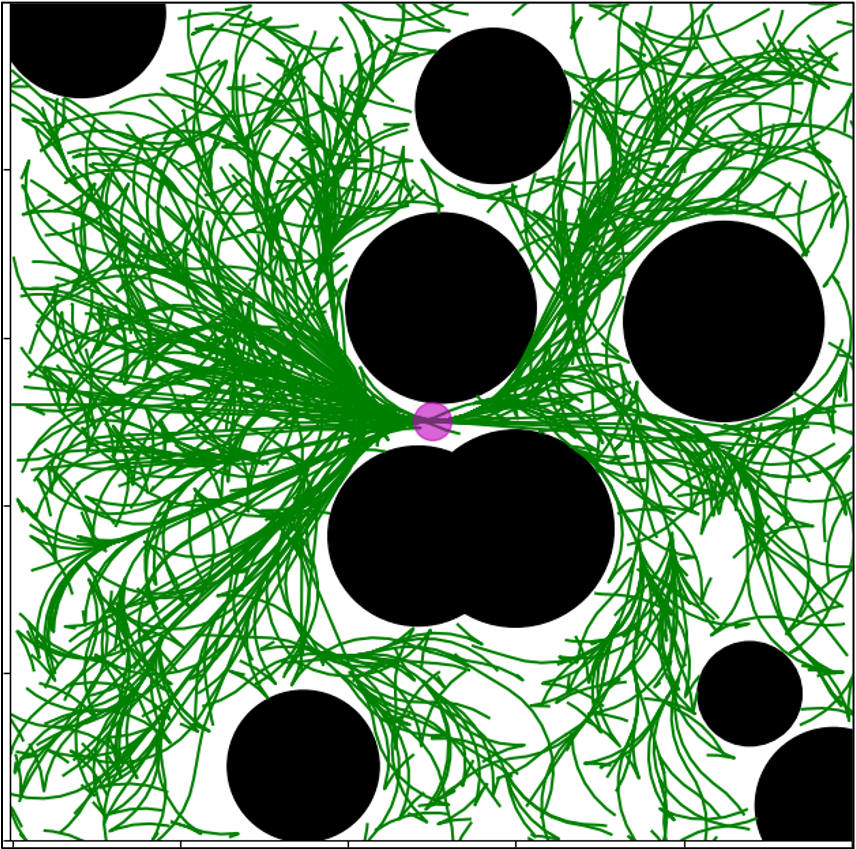}
\caption{} \label{fig:dataset_whole}
\end{subfigure}
\caption{Two phase tree expansion. (a) Initial collision-free branches by Reeds-Shepp curves. (b) Additionally generated branches by RRT* for exploring unreachable areas. (c) The entire tree structure.}
% \vspace{-3mm}
\label{fig:trees_for_dataset}
\end{figure}

This section presents the details of our methodology for adaptive sampling. 
%describes detail approaches for 
%learning of neural network for nonholonomic systems and adaptive sampling for training of network. 

The success of our approach relies on generating a dataset which captures high curvature regions on the motion planning manifold. 
Our dataset consists of workspace instances represented as point clouds of obstacles (or a workspace image) along with a list of triplets composed of source and destination configurations and the associated cost to go.

% The one sample data is composed of a set of tuples of pairwise configurations and a corresponding cost-to-go, and one workspace instance represented by a point cloud of collision regions. 

We start with generating random workspaces. For each workspace, we run RRT* with randomly initialized seed configurations until the generated trajectories cover the entire configuration space. % in cluttered environments. These multiple RRT* help to keep the diversity of dataset. 
%We generate a dataset of cost-to-go values using 
%traditional planners such as Reeds-Shepp curves and RRT* 
In addition, we apply optimal Reeds-Shepp curves for  paths to connect local nodes\cite{dubins1957curves,reeds1990optimal,karaman2011anytime}.

%As shown in Fig. \ref{fig:tree_cost}, we randomly sample two pairwise configurations ($s$ and $t$) in the same branch from the tree structure, and we can compute the cost easily from cost values in the tree structure.
%we assume that paths from the seed to leaf nodes in RRT* converge to the optimal paths by repeated rewiring,
%  the corresponding cost of these sampled configurations on the same branch of RRT* also has the minimum cost between these two configurations. 
We sample a pair of points ($s$, $t$) and the corresponding cost on the same branch of trees as shown in Fig. \ref{fig:tree_cost}.  Since a sub-path of an optimal path is also optimal and we assume that paths from RRT* converges to optimal paths asymptotically, the corresponding cost of these sampled configurations also has the minimum cost between them. 

To handle the angle wrapping, the orientation $\theta$ is mapped into $(\cos(\theta), \sin(\theta))$, and positions $x$ and $y$ are normalized to $[0,1]$.

Next, we describe the three factors critical to the performance of the dataset: diversity with respect to workspace, non-holonomic constraints, and maximizing total information.

\subsection{Two phase tree construction}
%. Since RRT* requires massive time-consuming rewiring routines to optimize tree structure, it goes slower as the number of nodes increases.

We  propose an efficient two phase tree construction method for RRT* as shown in Fig. \ref{fig:trees_for_dataset}. 
First, we generate an initial tree with collision-free Reeds-Shepp curves from a seed (Fig. \ref{fig:dataset_reeds_shepp}). Next, an RRT* starts exploring with this initial tree towards unexplored areas. Fig. \ref{fig:dataset_rrt_star} shows additionally explored branches in the second step and Fig. \ref{fig:dataset_whole} shows the constructed tree including all branches generated in the first and second steps.

Since branches in the initial tree by Reeds-Shepp curves (Fig. \ref{fig:dataset_reeds_shepp}) are optimal, the RRT* keeps the optimality by rewiring routines with only newly explored nodes and branches. This approach reduces the computation time of RRT* and the tree covers the whole configuration space effectively~(Fig. \ref{fig:dataset_whole}).

\subsection{Dataset for non-holonomic constraints}
 
\begin{figure}
\centering
\begin{subfigure}[b]{0.24\columnwidth}
\includegraphics[width=\textwidth]{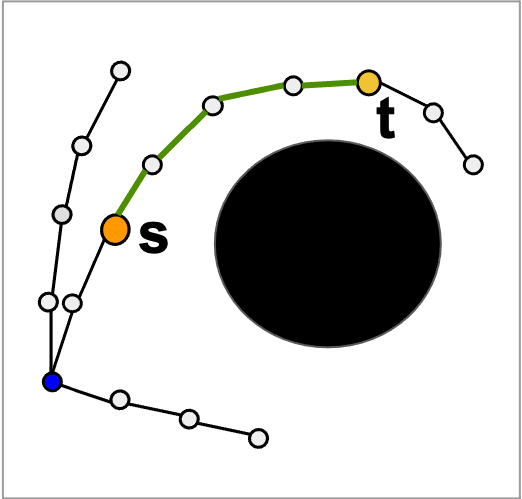}
\caption{} \label{fig:tree_cost}
\end{subfigure}
\begin{subfigure}[b]{0.24\columnwidth}
\includegraphics[width=\textwidth]{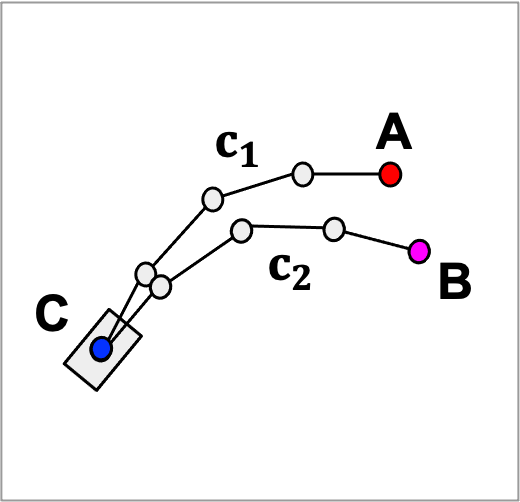}
\caption{} \label{fig:branch_cost}
\end{subfigure}
\begin{subfigure}[b]{0.20\columnwidth}
\includegraphics[width=\textwidth]{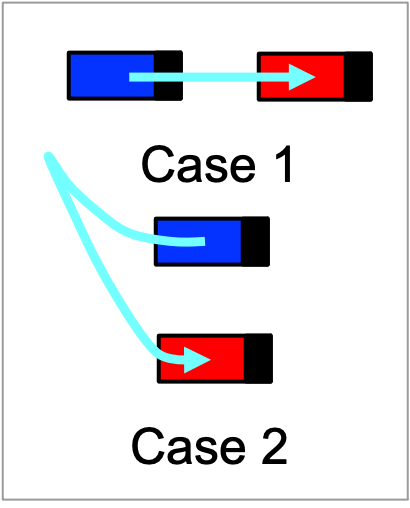}
\caption{} \label{fig:example_costs}
\end{subfigure}
\begin{subfigure}[b]{0.24\columnwidth}
\includegraphics[width=\textwidth]{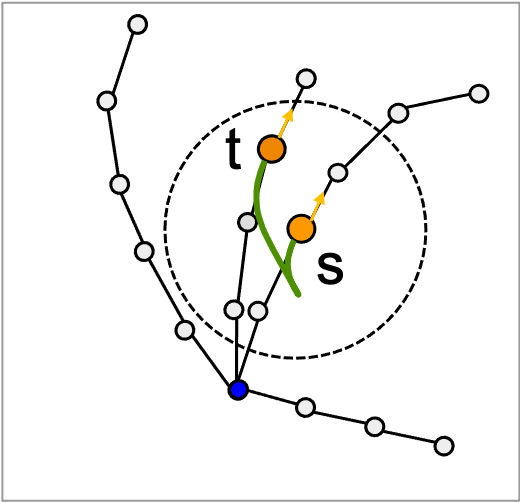}
\caption{} \label{fig:new_suggest_sample}
\end{subfigure}
\caption{ Sample generation. (a) We sample a pair of configurations ($s$, $t$) on the same branch of RRT* (b) The cost-to-go between $A$ and $B$ along  the tree does not reflect the min. cost (c) A start configuration is a blue rectangle, and a goal configuration is a red rectangle with orientations indicated by  black tips. Case 2 has a higher cost-to-go than Case 1 due to constraints. (d) To capture cases like (c), additional cost-to-go samples across vertices of RRT* within a threshold distance are generated. }
\label{fig:explain_limit_dataset}
% \vspace{-2mm}
\end{figure}

% \begin{figure}
% \centering
% \begin{subfigure}[b]{0.34\columnwidth}
% \includegraphics[width=\textwidth]{figs/tree_limit1.png}
% \caption{} \label{fig:tree_cost}
% \end{subfigure}
% \begin{subfigure}[b]{0.34\columnwidth}
% \includegraphics[width=\textwidth]{figs/tree_limit2.png}
% \caption{} \label{fig:branch_cost}
% \end{subfigure}
% \begin{subfigure}[b]{0.27\columnwidth}
% \includegraphics[width=\textwidth]{figs/tree_limit3.png}
% \caption{} \label{fig:example_costs}
% \end{subfigure}
% \caption{We randomly sample two pairwise configurations ($s$ and $t$) in the same branch from the tree structure, and we can compute the cost easily from cost values in the tree structure. However, we can compute the cost $c_1$ from $A$ to $C$, the cost $c_2$ from $A$ to $B$, but we don't know the cost between $A$ and $C$.The upper bound cost between between $A$ and $C$ is $c_1+c_2$. (c) Case 1 has a cost-to-go same as Euclidean distance between two configurations, Case 2 has longer cost-to-go compared to case 1 due to the constraint. Sample data from RRT* covers a type of case 1, but the type of case 2 is not well arisen during RRT*. }
% \label{fig:explain_limit_dataset}
% \end{figure}
Although we generate 10,000 workspace instances and 50,000 pairwise configurations for each instance by RRT* for diversity in various environments, we need additional effective data samples caused by non-holonomic constraints. Specifically, the dataset must capture paths across branches of the tree created in the previous section.

%we need an effective sampling strategy since nonholonomic constraints affect on cost-to-go values such that a parallel pose has longer geodesic distance than Euclidean distance. \vtxt{the previous sentence is hard to understand}
%\vtxt{maybe you can tie it to the manifold discussion in the intro and say that you are trying to sample high curvature regions}

In Fig. \ref{fig:branch_cost}, although the dataset from RRT* includes information of the cost $c_1$ from $A$ to $C$ and the cost $c_2$ from $A$ to $C$, it doesn't include the cost between $A$ and $C$. The costs $c_1$ and $c_2$ contain important information for collision avoidance. The cost between $A$ and $C$ is also informative due to non-holonomic constraints. In Fig. \ref{fig:example_costs}, Case 1 and Case 2 have the same Euclidean distance, but the non-holonomic constraint makes a difference in cost-to-go values of the two cases. Since RRT* expands the tree in the longitudinal direction to keep optimality by rewiring, it is difficult to include such samples in the dataset.

%since it represents features by nonholonomic constraint. The cost between two configurations is sometimes large although they are close in Euclidean space as shown in Case 2 of Fig \ref{fig:example_costs}. 

%\vtxt{can you show an example of this in the figure? you can pick a random pair (s and t from each branch) and draw a cartoon reeds-shepp curve and explain it in the caption}
Therefore, we additionally sample cost-to-go across vertices of RRT* computed by Reeds-Shepp curves as shown in Fig. \ref{fig:new_suggest_sample}. We sample pairs of vertices within a threshold distance $\alpha * \rho$, where $\alpha$ is a constant ($\alpha = 1.5$ in this paper) and $\rho$ is a lower bounded turning radius by the non-holonomic constraint. In the region outside of the turning radius, the cost for collision avoidance is more dominant as shown in trained cost-to-go images (Fig. \ref{fig:two_phase_cost}). 
%The trained network with this suggested compensation captures features of nonholonomic constraints as shown in Fig. \ref{fig:two_phase_cost}.

%Because of this reason, the global cost distribution is appropriate, the local trajectory can be inaccurate around the goal configuration. The second difficulty is that there is no optimal method for generating cost-to-go values for training. We suggest an adaptive sampling for dataset generation method. 
 
\subsection{Adaptive sampling based on diversity of gradient}

\begin{figure}
\centering
\begin{subfigure}[b]{0.39\columnwidth}
\includegraphics[width=\textwidth]{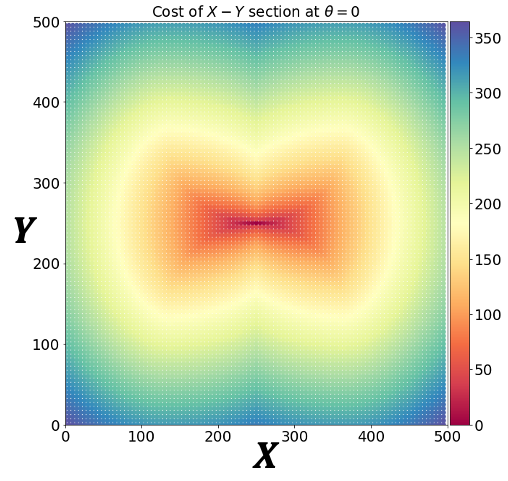}
\caption{} \label{fig:cost_x_y}
\end{subfigure}
\begin{subfigure}[b]{0.39\columnwidth}
\includegraphics[width=\textwidth]{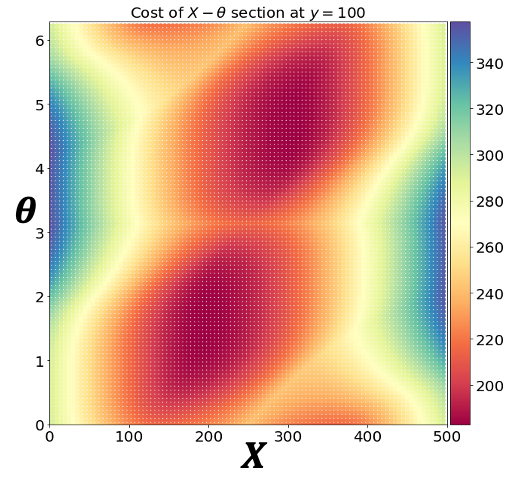}
\caption{} \label{fig:cost_x_theta}
\end{subfigure}
\begin{subfigure}[b]{0.39\columnwidth}
\includegraphics[width=\textwidth]{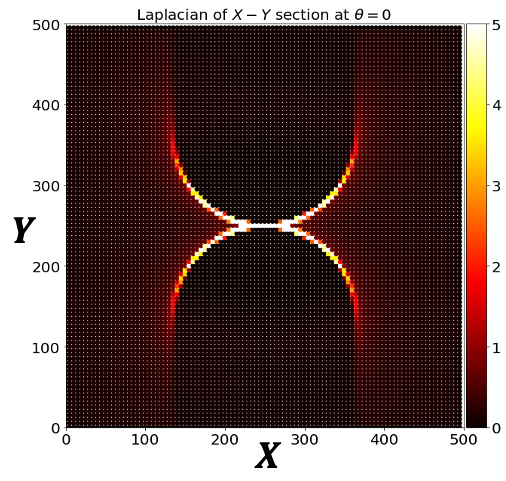}
\caption{} \label{fig:laplacian_x_y}
\end{subfigure}
\begin{subfigure}[b]{0.39\columnwidth}
\includegraphics[width=\textwidth]{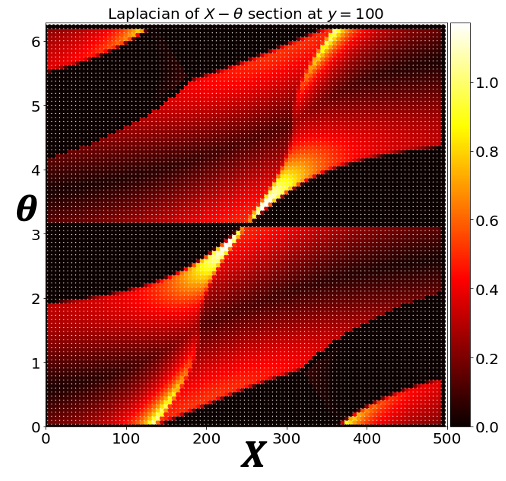}
\caption{} \label{fig:laplacian_x_theta}
\end{subfigure}
\caption{Visualization of cost-to-go for the goal configuration at $(250, 250, 0)$ (a) $X-Y$ plane with $\theta = 0$ (b) $X-\theta$ plane with $y = 100$ (c) Laplacian of cost-to-go shown in (a) (d) Laplacian of cost-to-go shown in (b) (Best viewed in color).}
\label{fig:cost_laplacian}
\vspace{-3mm}
\end{figure}

%\vtxt{add an intro paragraph where you say rather than uniformly sampling configurations, we use a weighted sampling scheme to boost the likelihood of sampling regions with high curvature. And then describe the method}

%and training data important in training of the network.
%Several approaches are suggested to learn importance of samples along with training the network.

As described in \cite{katharopoulos2018not,fan2017learning}, all data samples are not equally critical for the neural network performance. A good dataset can accelerate the training and impact the performance of the network.  Uniform sampling can cover the whole space, but it leads to including redundant data due to biased data distribution. Therefore, we need an additional effective sampling strategy since non-holonomic constraints ``curve" the manifold of cost-to-go as shown in Fig. \ref{fig:high_curvature}. We can see that the cost-to-go is distinct from Euclidean distance due to the non-holonomic constraint even without an obstacle in Fig. \ref{fig:cost_x_y} and \ref{fig:cost_x_theta}.
Therefore, we need to consider the curvature of \ctog{}, since uniform sampling is difficult to represent the manifold of cost-to-go.

\begin{figure}[t]
\centering
\begin{subfigure}[b]{0.49\columnwidth}
\includegraphics[width=\textwidth]{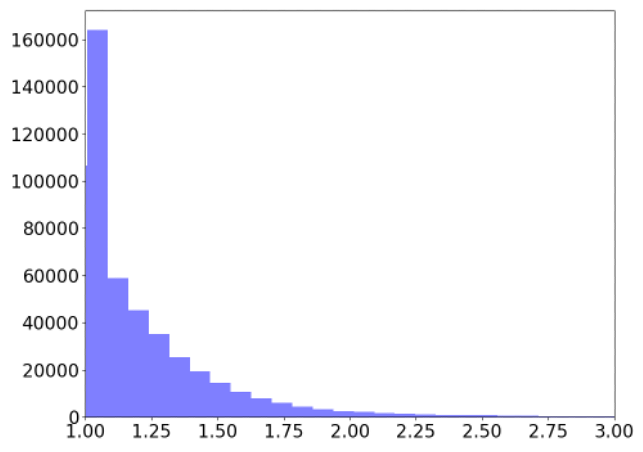}
\caption{Uniform sampling} \label{fig:histogram_uniform}
\end{subfigure}
\begin{subfigure}[b]{0.49\columnwidth}
\includegraphics[width=\textwidth]{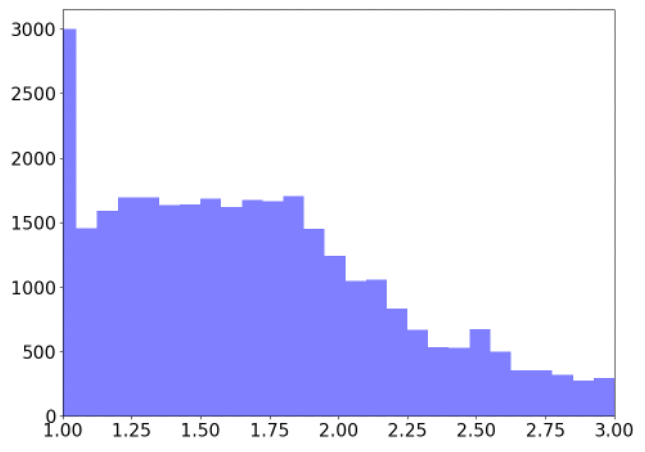}
\caption{Adaptive sampling} \label{fig:histogram_laplacian}
\end{subfigure}
\caption{
Histogram of the ratio of true cost to Euclidean distance. High curvature regions would have a higher ratio. Uniform sampling heavily samples regions resembling Euclidean distance (ratio$\approx$1), whereas adaptive sampling distributes the samples more evenly. }
\label{fig:explain_histogram_dataset}
\vspace{1mm}
\end{figure}

We consider the curvature of cost-to-go by calculating its Laplacian, shown in Figs. \ref{fig:laplacian_x_y} and \ref{fig:laplacian_x_theta} without obstacles. We need dense sampling around high Laplacian areas since high Laplacian means high curvature. We can see that regions around goal configuration have higher Laplacian. In environments with obstacles, since it is hard to compute the Laplacian directly, we compute a ratio of the cost-to-go gradient to the Euclidean distance gradient. Practically, if this ratio is high, we sample more with high probability. Fig. \ref{fig:explain_histogram_dataset} shows ratio of samples histograms based on uniform sampling and adaptive sampling. The ratio of gradients by uniform sampling is strongly biased to 1 in Fig. \ref{fig:histogram_uniform}. A ratio of 1 means the gradient of samples are the same as the gradient of Euclidean distance, and samples with higher ratio are more informative.

% \begin{itemize}
%     \item Sampling based on Laplacian.. 
%     \item Figures shows laplacian in open space
%     \item In obstacle environment, it is hard to compute Laplacian directly. we compare with Eclidean distance and compute divergen
%     \item Need to show histograms before and after
%     \item  Most informative
%     \item scalar valued function of X and Y
%     \item It is kind of like a second derivative
%     \item The divergence of the gradient of the cost-to-go function
%     \item This is the sense in which it is a second derivative
% \end{itemize}

\section{Neural Network Architecture}
In this paper, for evaluation of suggested methods to learn cost-to-go functions of non-holonomic systems, we choose higher order function network (HOF) architecture proposed in~\cite{mitchell2019higher} and applied to motion planning in our previous work~\cite{huh2020cost}.
%Since the cost-to-go function $f_w$ is changed by the workspace, HOF is appropriate for learning of cost-to-go in various environmental conditions. 

\begin{figure}[t]
\centering
\includegraphics[width=0.91\columnwidth]{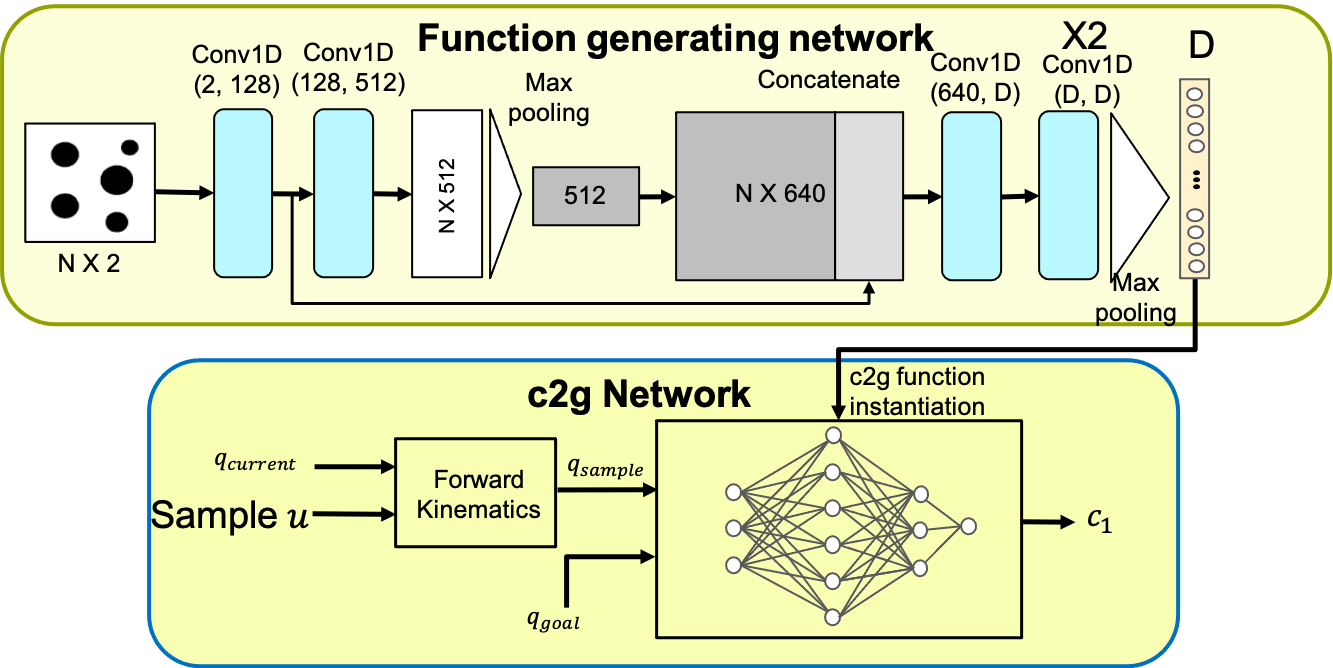}
\caption{Non-holonomic c2g-function network architecture}
\label{fig:total_network}
\end{figure}

The version we use for the present work, which we call the  non-holonomic c2g-function network, is composed of two sub-networks; one sub-network is a function generating network by the HOF network and the other sub-network is the c2g-function (Fig. \ref{fig:total_network}). 
In this paper, we use PointNet \cite{qi2017pointnet} as the encoder for the function generating network where we sample the obstacles in the workspace $\mathcal{W}$. The c2g-function network has three perceptron layers which have 256 neurons and a ReLU activation layer at each layers.

Once the network is trained, the function generating network encodes a point cloud of the workspace and outputs parameters of the c2g function network. The c2g-function network is instantiated with these output parameters.
For the input of c2g-function network, sample configurations are concatenated with the goal configuration. Before the input layer of the c2g-function network, there is forward kinematics layer which computes sample configurations from sample control inputs.

% \subsection{Neural Network Architecture}

\subsection{Training}

From the training dataset, we sample 50,000 pairwise configurations and corresponding ground-truth costs in one workspace instance. We input a workspace point cloud into the function generating network and construct the c2g-function network with the output from the function generating network. The constructed c2g-function network inputs concatenated pairwise configurations and it outputs predicted costs for sample configurations. Since the parameters of c2g-function networks are outputs of the function generating network, the Mean Squared Errors (MSE) between predicted costs and ground truth costs are backpropagated to the function generating network, which learns to output parameters of cost-to-go functions correctly to reduce MSE. 

    % \item Difference from previous c2g-HOF
    %     \begin{itemize}
    %         \item Same : We present a Higher Order Function (HOF) network which generates a continuous cost-to-go function of two pair points over the entire configuration space based on workspace input. The trajectory can be generated by following the gradient of cost-to-go
    %         \item Difference: It is very difficult to learn the network with uniform samplings obtained by a planner such as RRT*. We suggest an efficient sampling method for dataset generation for c2g-HOF.
    %     \end{itemize}
        
\subsection{Trajectory Generation}

We use cost-to-go function networks for path planning of non-holonomic systems. For the trajectory generation, first, control inputs are sampled and then a forward kinematics layer computes the next sample configurations with the current configuration and sampled control inputs. The next sample configurations are concatenated with the goal configuration and they are the input of the c2g-function network which outputs cost-values. 

The network outputs a set of traversal costs from sample configurations to the goal configuration. It then removes infeasible control inputs such as collision with obstacles. The planner chooses the control input which has the minimum cost, and the configuration corresponding to this control input can be the next configuration. This routine repeats until the goal configuration.
             
The planner’s stopping criteria for reaching the goal configuration is as follows: a) the current cost value is less than a threshold or b) the new configuration position has a smaller difference than a threshold and the orientation is less than a threshold. The sequence of current configurations can be the waypoint of the trajectories.
            
        % \begin{itemize}
        %     \item First it sample control inputs. 
        %     \item Note that we assume that it has forward and backward motion input with a lower-bounded of turning radius.
        %     \item It computes sample configurations from current configuration with sampled control inputs based on forward kinematics. 
        %     \item The network outputs a set of traverse costs from sample configurations to the goal configuration.
        %     \item It removes infeasible control inputs such as collision with obstacles.
        %     \item We choose the control input which has the minimum cost, and the configuration corresponding to this control input for the waypoint for the trajectory.
        %     \item It repeats until the goal configuration.
        %     \item For the Criteria for arriving the goal configuration. It stops if current cost value is less than threshold. or It stops if the new configuration is less than some threshold of position and angle heading (no metric for both position and angle), small position error makes large error. In the less than some boundary, we use Reeds-Shepp curve for perfect trajectory.  
        % \end{itemize}    

    % The library implements the analytic Reeds Shepp path between two SE2 configurations. Reeds Sheep path[^1] is defined as the shortest traveling path of the Reeds-Shepp Car, a car that can go both forward and backward with a constrained turning radius. 

\section{Results}
This section presents experiments comparing our approach to existing planners in terms of trajectory length and computation time. In addition, to verify our sampling methodology, we show results obtained by trained networks with different sampling datasets, and verify the reliability of our approaches via a mobile robot simulator.

\subsection{Learning of \ctog{}}
  
 %\vtxt{add a short introductory/overview paragraph}
\begin{figure}
\centering
\begin{subfigure}[b]{0.43\columnwidth}
\includegraphics[width=\textwidth]{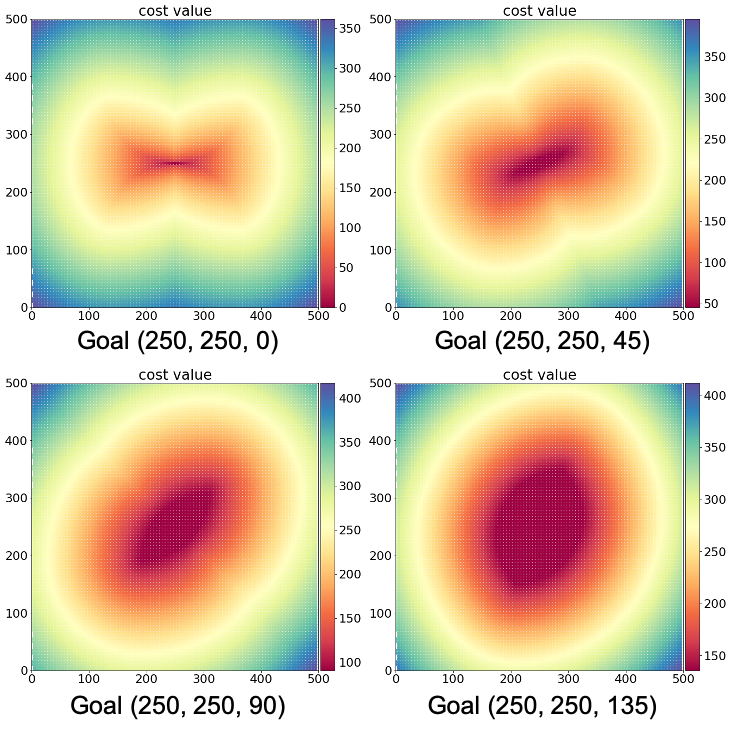}
\caption{GT by Reeds-Shepp} \label{fig:cost_gt}
\end{subfigure}
\begin{subfigure}[b]{0.43\columnwidth}
\includegraphics[width=\textwidth]{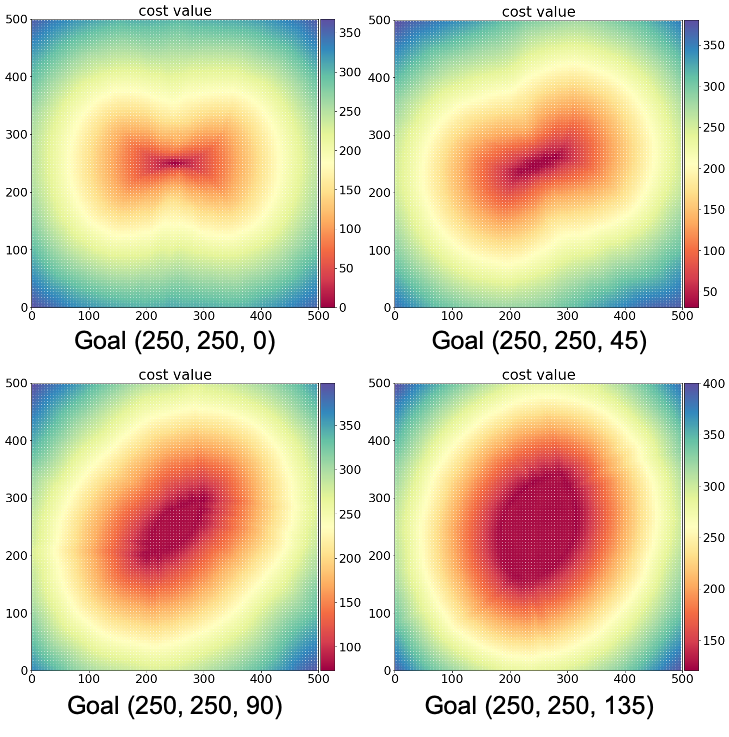}
\caption{Prediction by network} \label{fig:cost_network}
\end{subfigure}
\caption{Cost-to-go for various goal configurations in environments with no obstacles (a) ground truth (Reeds-Shepp curves) (b) the prediction by the trained network. For visualization, figures show cross-sections at $\theta = 0$ (Best viewed in color).}
\label{fig:compare_cost}
\end{figure}

%In order to evaluate predicted \ctog{} by the trained network with suggested dataset generation, we compare with optimal ground truth \ctog{}. 
Figure \ref{fig:compare_cost} shows a comparison of optimal ground truth \ctog{} by Reeds-Shepp curves and predicted \ctog{} by the network in an open environment with no obstacles. 
For the ground truth (Fig. \ref{fig:cost_gt}), we compute lengths of all Reeds-Shepp curves between grid points of $(x,y, \theta)$ and several different goal configurations. For the predicted \ctog{} (Fig. \ref{fig:cost_network}), the trained network inputs the same grid configurations concatenated with a goal configuration.
For visualization, Fig. \ref{fig:compare_cost} shows cross-sections of the \ctog{} at $\theta = 0$ for multiple goal configurations. Fig. \ref{fig:compare_cost} shows that the trained network predicts \ctog{} accurately over the entire continuous \cspace{}. In addition, we can see how the cost changes with changing goal configuration. Note that the prediction of \ctog{} for continuous configuration inputs is a strong advantage of learning of \ctog{} function network. 

\begin{figure}
\centering
\begin{subfigure}[b]{0.45\columnwidth}
\includegraphics[width=\textwidth]{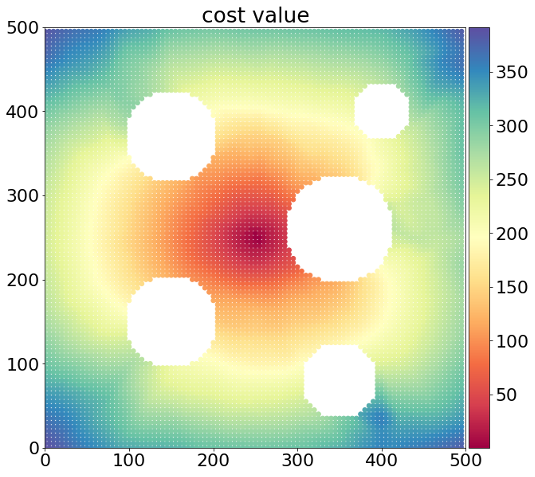}
\caption{Uniform sampling} \label{fig:rrt_only_cost}
\end{subfigure}
\begin{subfigure}[b]{0.45\columnwidth}
\includegraphics[width=\textwidth]{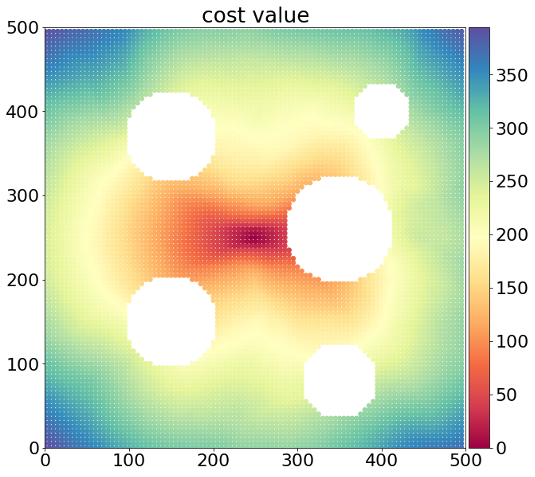}
\caption{Adaptive sampling} \label{fig:two_cost}
\end{subfigure}
\caption{Predicted cost-to-go by trained networks with two different datasets (a) uniform sampling dataset (b) adaptive sampling dataset. Figures show cross-sections at $\theta = 0$ for the goal configuration at $(250, 250, 0)$. Adaptive sampling keeps high cost values behind obstacles and captures gradients well around the goal configuration (Best viewed in color).}
\label{fig:two_phase_cost}
\end{figure}

In cluttered environments, we compare our approach against a uniform sampling method. Fig. \ref{fig:two_phase_cost} shows comparison results by showing cross-sections of \ctog{} at $\theta  = 0$ for the goal configuration at $(250, 250, 0)$. 
In the training result with uniform sampling (Fig. \ref{fig:rrt_only_cost}), it has high \ctog{} behind obstacles to avoid obstacles. Note the lack of features around goal configuration. In contrast, the training result with the adaptive sampling (Fig. \ref{fig:two_cost}) keeps high cost values behind obstacles and captures the changes of the gradient around the goal configuration. 
 Since the changes of gradient are caused by the non-holonomic constraint, gradients around goal configurations are critical information to arrive at the goal configuration precisely.

%  by networks trained with two different datasets.

\subsection{Trajectory generation}
\label{sec:traj_generation}

\begin{figure}[t]
\centering
\includegraphics[width=\columnwidth]{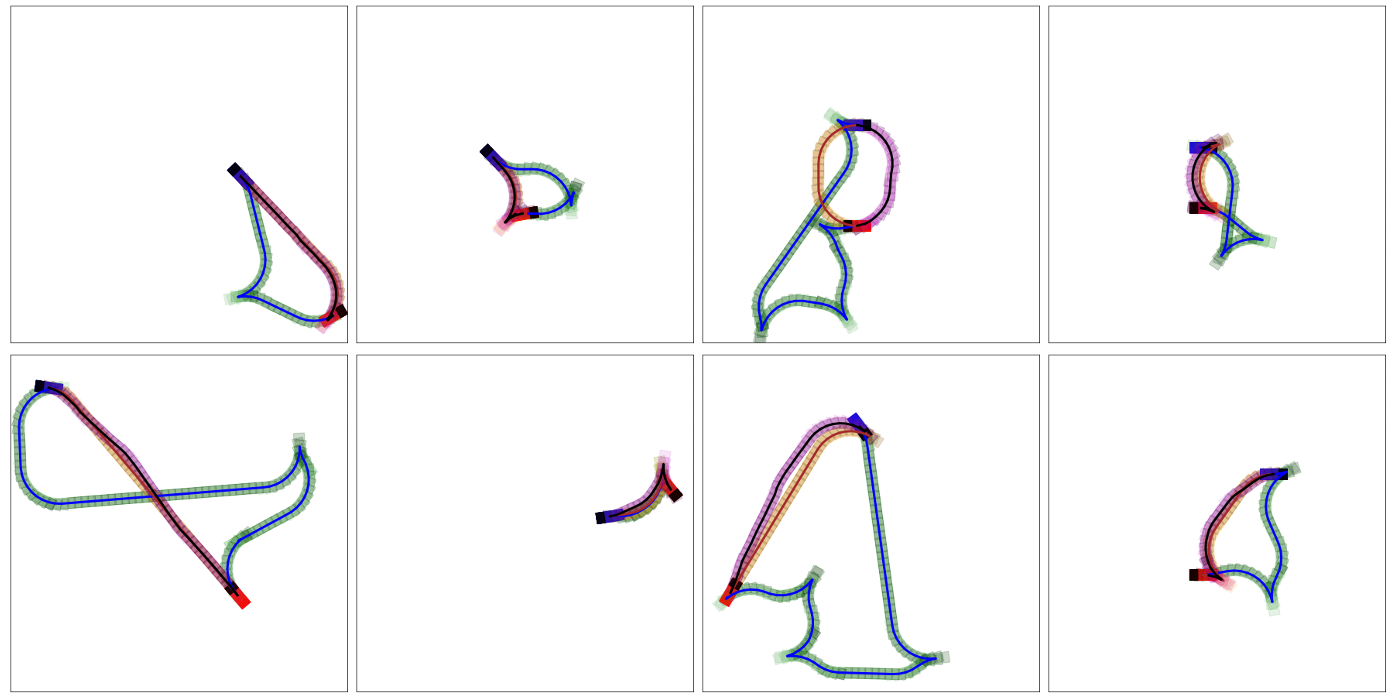}
\caption{Trajectories in the open space by RRT (blue), RRT* (orange), and \ctogplanner{} (magenta).}
\label{fig:traj_open_result}
\end{figure}

\begin{figure}[t]
\centering
\includegraphics[width=\columnwidth]{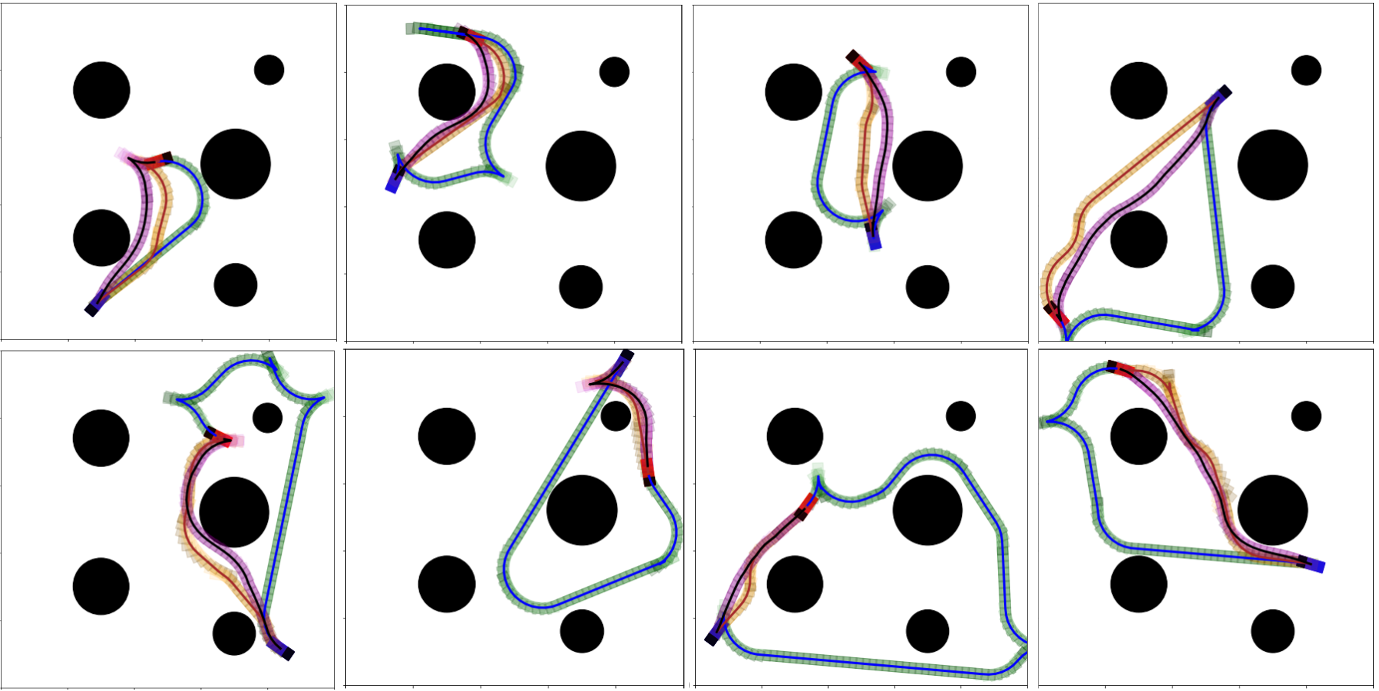}
\caption{Trajectories for various start and goal configurations in an environment with obstacles by RRT (blue), Reeds-Shepp curves (orange), and \ctogplanner{} (magenta) based on cost-to-go of Fig. \ref{fig:two_cost}. }
\label{fig:traj_clutter_result}
\end{figure}

\begin{figure}[t]
\centering
\includegraphics[width=\columnwidth]{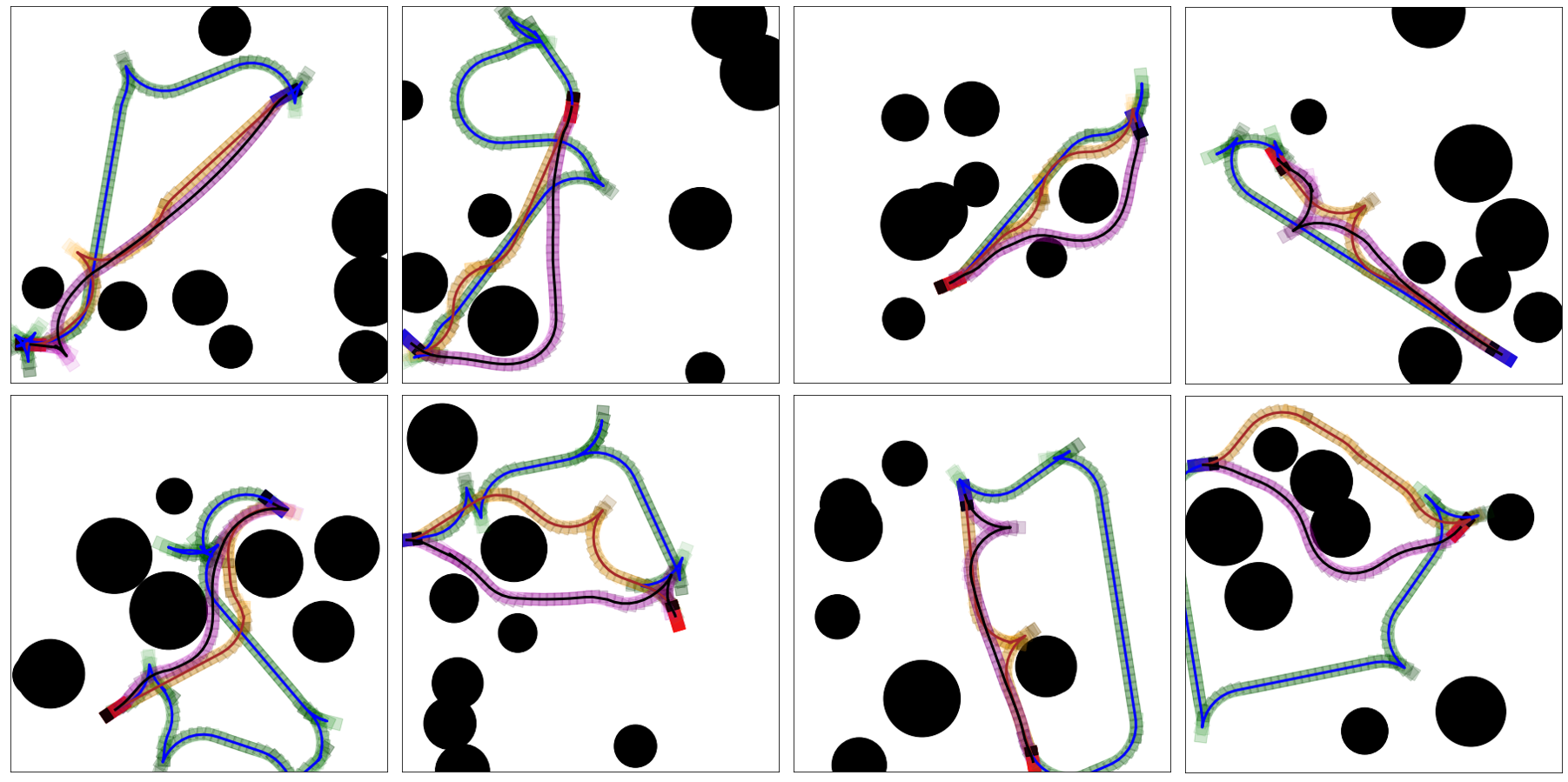}
\caption{Trajectories for various start and goal configurations in various cluttered environments by RRT (blue), RRT* (orange), and \ctogplanner{} (magenta).}
\label{fig:traj_clutter_changed_result}
\end{figure}

\begin{figure}[t]
\centering
\begin{subfigure}[b]{0.47\columnwidth}
\includegraphics[width=\textwidth]{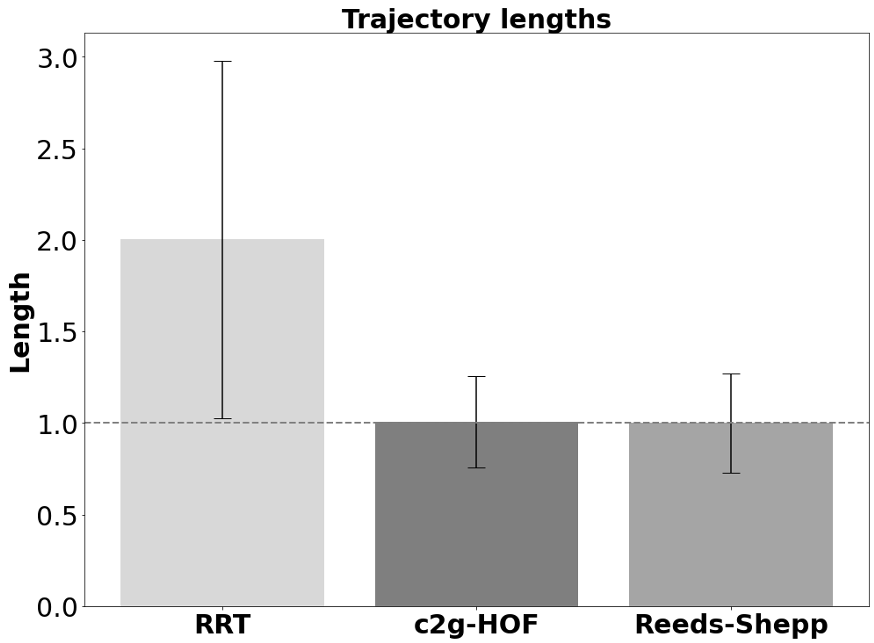}
\caption{No obstacle} \label{fig:result_open_length}
\end{subfigure}
\begin{subfigure}[b]{0.47\columnwidth}
\includegraphics[width=\textwidth]{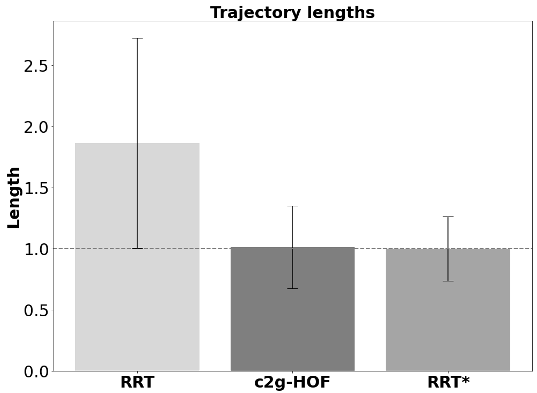}
\caption{Cluttered environment} \label{fig:result_clutter_length}
\end{subfigure}
\caption{Average trajectory length (a) trajectories in environments with no obstacles in Fig. \ref{fig:traj_open_result} (b) trajectories in various cluttered environments in Fig. \ref{fig:traj_clutter_changed_result}. The trajectory length is normalized by the average trajectory length of optimal Reeds-Shepp for the first case  and RRT* for the latter.}
\label{fig:result_gen3}
\vspace{0mm}
\end{figure}

\begin{figure}[t]
\centering
\begin{subfigure}[b]{0.47\columnwidth}
\includegraphics[width=\textwidth]{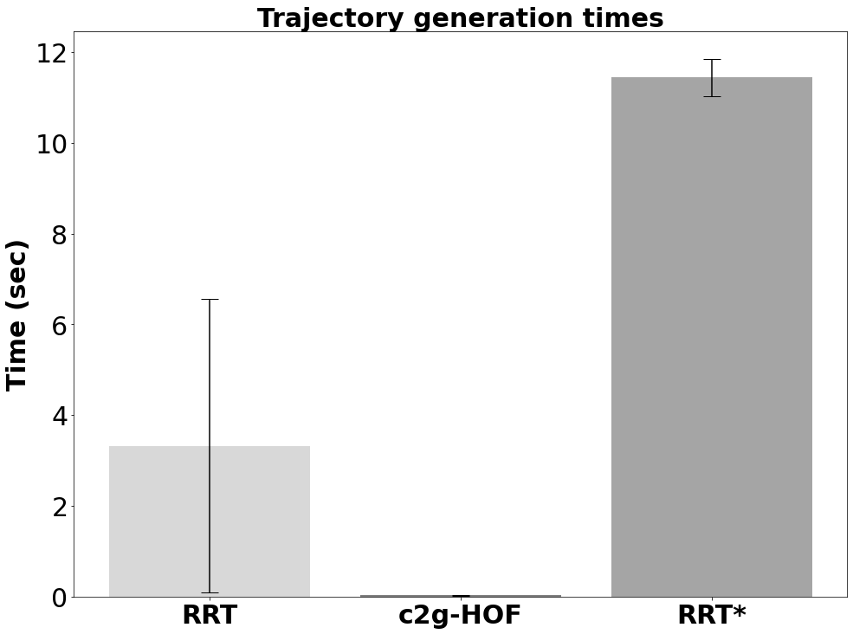}
\caption{Time of Fig. \ref{fig:traj_clutter_result}} \label{fig:result_computation_time_5obs}
\end{subfigure}
\begin{subfigure}[b]{0.47\columnwidth}
\includegraphics[width=\textwidth]{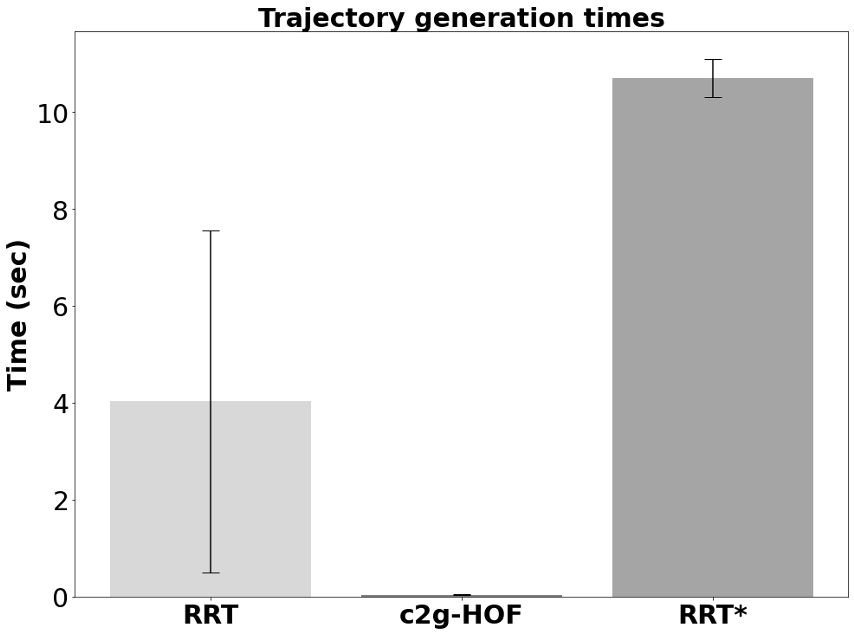}
\caption{Time of Fig. \ref{fig:traj_clutter_changed_result} } \label{fig:result_computation_time_clutter}
\end{subfigure}
\label{fig:result_computation_time}
\caption{ Average trajectory generation time. Left to right: RRT, ours with adaptive sampling, RRT*}
\end{figure}

We compare the performance of the \ctogplanner{} with other approaches, such as RRT, RRT*, and Reeds-Shepp methods in terms of trajectory length and computational time. In comparison, the RRT* planner is the same as the method used for data generation. 
%It generates initial Reeds-Shepp curves and the tree extends towards unreachable points based on initial Reeds-Shepp curves. 
Although RRT* requires heavy computation, we compare ours with RRT* in terms of trajectory length. 
In addition, we apply 3\% goal configuration biased sampling for RRT to accelerate it reaching the goal configuration.
Reeds-Shepp curves are used for local planners of RRT and RRT* to connect between nodes. In the environment with no obstacle, we compare ours with optimal Reeds-Shepp curves. To speed up these comparison methods, we use Python API library of C++ codes by Open Motion Planning Library (OMPL) \cite{sucan2012the-open-motion-planning-library} for Reeds-Shepp curves \cite{Reeds_Shepp_code}.

Fig. \ref{fig:traj_open_result} shows comparison results of trajectories by ours, RRT, and Reeds-Shepp curves. 
A start configuration is a blue rectangle, and a goal configuration is a red rectangle with orientations indicated by a black tip. Ours generates near-optimal trajectories which have almost the same length as the optimal Reeds-Shepp curves as shown in Fig. \ref{fig:traj_open_result}. The average trajectory length of ours is only 0.1\% longer than the average trajectory length of optimal Reeds-Shepp curves and 50\% less than the average trajectory length of RRT planner as shown in Fig. \ref{fig:result_open_length}.

Fig. \ref{fig:traj_clutter_result} shows generated trajectories with \ctog{} shown in Fig. \ref{fig:two_cost} for multiple random start and goal configurations. Test start configurations, goal configurations, and test environments are not included in the training dataset. The trajectory generations are compared with RRT and RRT*. We can see that the trajectories by ours have shorter distances while avoiding obstacles compared to the slow RRT* planner or the inefficient RRT trajectories. Fig. \ref{fig:result_computation_time_5obs} shows that computation time for ours (0.029 seconds) is two orders of magnitude faster than RRT (3.32 seconds) and RRT* (11.4 seconds).

We conduct quantitative comparison in various cluttered environments in terms of averages of computation time and trajectory length. Test start and goal configurations, and test environments are not included in the training dataset.  Fig. \ref{fig:traj_clutter_changed_result} shows generated trajectories, and the average length of ours is only 1.1\% longer than the average length of RRT* trajectories and 46\% shorter than the average length of RRT trajectories (Fig. \ref{fig:result_clutter_length}). The average computation time is only 45ms compared to other approaches taking several seconds (Fig. \ref{fig:result_computation_time_clutter}). We emphasize that ours has similar trajectory lengths to time-consuming RRT* and it takes less than 0.1 seconds for computation. Moreover, during trajectory generation, it does not need a local planner such as Reeds-Shepp curves which are used for connecting vertices in RRT and RRT*, while most planners require a local planner for satisfying non-holonomic constraints along with a global trajectory. 

The success rate of our approach is 100\% in cases of Figs. \ref{fig:traj_open_result} and \ref{fig:traj_clutter_result}, and 98.5\% in the case of Fig. \ref{fig:traj_clutter_changed_result}. 
%0.039, 4.03 seconds, 10.71 seconds

\begin{figure}[t]
\centering
\begin{subfigure}[b]{0.47\columnwidth}
\includegraphics[width=\textwidth]{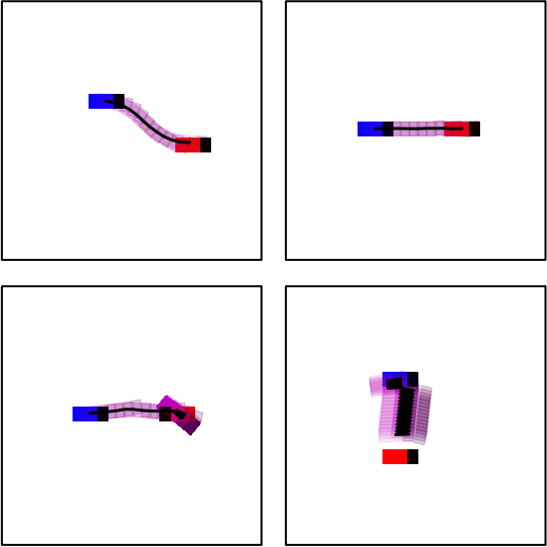}
\caption{Uniform dataset} \label{fig:c2g_failure}
\end{subfigure}
\begin{subfigure}[b]{0.47\columnwidth}
\includegraphics[width=\textwidth]{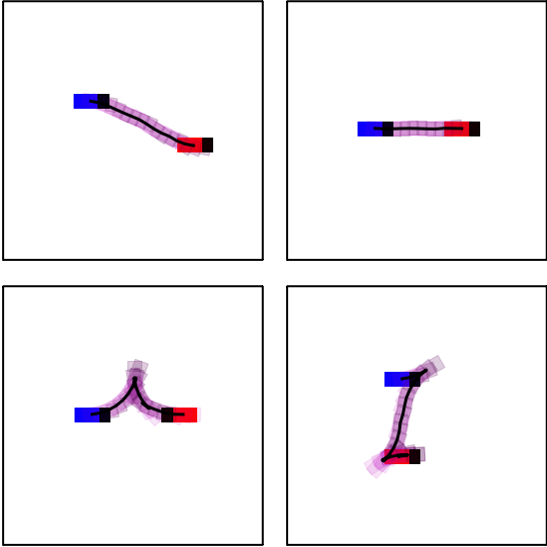}
\caption{Adaptive dataset} \label{fig:nonhol_c2g_success}
\end{subfigure}
\caption{Training with uniform (a) vs. adaptive sampling (b). Start configurations are marked blue and goal configurations (red) with the black part showing the orientation. As the curvature increases, uniform-sampling instances become less accurate.} 
\label{fig:comparison_c2g_HOF}
\vspace{0mm}
\end{figure}

In order to verify the effectiveness of the suggested sampling method, we test several trajectory planning tasks as shown in Fig. \ref{fig:nonhol_c2g_success}. Fig. \ref{fig:c2g_failure} shows trajectories generated by the network trained with the uniform sampling dataset and Fig. \ref{fig:nonhol_c2g_success} shows trajectories generated by the network trained with the suggested sampling dataset. The top two figures show planning tasks in which cost-to-go values are similar to Euclidean distance from a start configuration (blue) to a goal configuration (red). In these tasks, both networks generate trajectories appropriately. In the bottom-left task, the relative angle between the start and the goal configurations is $\pi$ and has a small relative distance in $x$-$y$ plane. In this case, the trajectory based on the network trained with the uniform sampling dataset is just heading to the goal position without considering the orientation. However, the network trained with the suggested sampling dataset changes the orientation in the middle of the trajectory and arrives at the goal configuration correctly. In the bottom-right task, the goal configuration is parallel to the start configuration. The trajectory based on the network trained with the uniform sampling dataset and repeats inefficient forward and backward movements from the start configuration to the goal configuration. The network trained with the suggested sampling dataset generates a near-optimal path trajectory.

% Compare with RRT and RRT*
% RRT & RRT*
% In order to improve the performance, RRT& RRT* uses Reeds-Shepp (C++ library by OMPL) for local planner, and 5% goal configuration biased sampling
% For the optimality and performance, RRT* uses initial collision-free Reeds-Shepp curves from start points and uses rewiring for avoiding obstacles
% 1.1% more length compared to RRT* in the environment with obstacles 
% (100 random trajectories)

\begin{figure}[t]
\centering
\begin{subfigure}[b]{0.37\columnwidth}
\includegraphics[width=\textwidth]{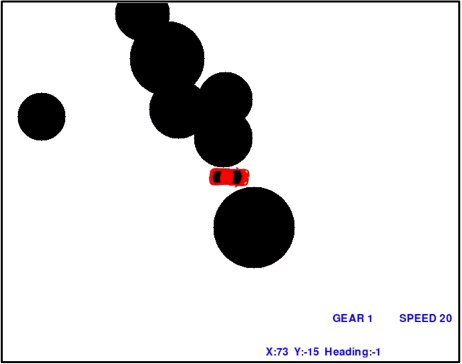}
\caption{Simulation} \label{fig:sim_env}
\end{subfigure}
\begin{subfigure}[b]{0.29\columnwidth}
\includegraphics[width=\textwidth]{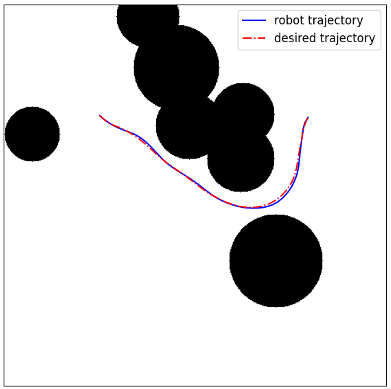}
\caption{Trajectory} \label{fig:sim_traj}
\end{subfigure}
\begin{subfigure}[b]{0.3\columnwidth}
\includegraphics[width=\textwidth]{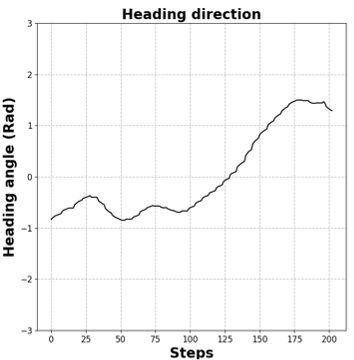}
\caption{Orientation} \label{fig:sim_heading}
\end{subfigure}
\caption{(a) Simulation environment (b) the mobile robot trajectory (blue) tracking a  trajectory (red) from \ctogplanner{} (c) Orientation of the mobile robot. We can see that the robot follows the desired trajectory smoothly without an abrupt change of states.} 
\label{fig:sim_results}
\vspace{0mm}
\end{figure}

\subsection{Simulation}
In order to demonstrate the reliability of the generated trajectories, we conduct a simulation in test environments with a car with more realistic dynamics (Fig. \ref{fig:sim_env}). In this simulation, the minimum turning radius of the mobile robot is $5.88 m$ with $4.5m/s^2$ acceleration. The maximum speed is $11.25 m/s$ and the mobile robot needs to stop and move again whenever it changes the direction. We apply a look-ahead controller for tracking the desired trajectory generated by the \ctogplanner{} to minimize the lateral error. 
As shown in Figs. \ref{fig:sim_traj} and \ref{fig:sim_heading}, the mobile robot follows the desired trajectory well and generates smooth changes of orientation. Compared to Dubins or Reeds-Shepp curves which generate only minimum turning curves or straight lines, ours generates continuous steering control inputs. Therefore, the trajectories generated by ours are more reliable for real systems. 

\section{Conclusion}
\vspace{-0.7mm}
This paper focused on learning based approaches for continuous cost-to-go function generation for non-holonomic systems. The high curvature regions of the optimal path manifold of non-holonomic systems make it difficult to represent them with uniform samples. 
We suggested an adaptive sampling approach based on the diversity of the gradient of the cost-to-go and showed that it  leads to a successful representation of cost-to-go in C-space. %We also showed that  uniform sampling makes learning slow or fails to represent cost-to-go.

To show the effectiveness of our approach, we focused on the Reeds-Shepp car. We used a similar architecture to our recent work~\cite{huh2020cost} where we presented cost-to-go generation networks for holonomic systems. We showed that the same network with uniform sampling is not effective 
for the Reeds-Shepp car. However, once trained with the presented adaptive sampling method, it generates the  parameters of an accurate cost-to-go function network almost instantly.
The planning based on the learned cost-to-go function network takes less than 0.1 seconds to generate near-optimal trajectories while other sampling-based planners take a few seconds. We also verified the reliability of generated trajectories in the mobile robot simulator. For future work, we plan to extend the approach to manipulators with smoothness constraints where sampling efficiency is critical due to the high dimension of the C-space.

%we will extend this framework to higher dimensional space problems with constraints such as a manipulator with an orientation constraint of end-effector.
%We will release dataset and code soon with the simulator in the final version.

\addtolength{\textheight}{-6.5cm}  % This command serves to balance the column lengths

\bibliographystyle{IEEEtran}
\bibliography{references}

%\listoftodos

\end{document}